\documentclass[10pt,twocolumn,letterpaper]{article}

\usepackage{wacv}
\usepackage{times}
\usepackage{graphicx}
\usepackage{amsmath}
\usepackage{amssymb}
\usepackage{algorithm,algorithmicx}
\usepackage[noend]{algpseudocode}
\usepackage{hyperref}
\hypersetup{colorlinks=true,urlcolor=blue,citecolor=blue,colorlinks,bookmarks=false,citecolor=blue,linkcolor=blue}
\makeatletter
\def\BState{\State\hskip-\ALG@thistlm}
\makeatother

\graphicspath{{grf/}}

\newcommand{\I}{\ensuremath{\mathbf{I}}}

\newcommand{\calR}{\ensuremath{\mathcal{R}}}

{%
\begin{list}{#1}{
\vspace{-\topsep}
\vspace{-\partopsep}
\setlength{\itemindent}{0cm}
\setlength{\rightmargin}{0cm}
\setlength{\listparindent}{0cm}
\settowidth{\labelwidth}{#1}
\setlength{\leftmargin}{\labelwidth}
\addtolength{\leftmargin}{\labelsep}
\setlength{\itemsep}{0cm}
}%
}%
{%
\end{list}
\vspace{-\topsep}
\vspace{-\partopsep}
}

{\begin{enumerate}%
}%
{\end{enumerate}}

\wacvfinalcopy 

\def\wacvPaperID{634} 

\ifwacvfinal\fi
\begin{document}

\title{Ventral-Dorsal Neural Networks: Object Detection via Selective Attention}

\author{Mohammad K. Ebrahimpour \\ 
UC Merced\\
{\tt\small mebrahimpour@ucmerced.edu}
\and
Jiayun Li\\
UCLA\\
{\tt\small jiayunli@ucla.edu}
\and
Yen-Yun Yu\\
Ancestry.com\\
{\tt\small yyu@ancestry.com}
\and
Jackson L. Reese\\
Ancestry.com\\
{\tt\small jreese@ancestry.com}
\and
Azadeh Moghtaderi\\
Ancestry.com\\
{\tt\small amoghtaderi@ancestry.com}
\and
Ming-Hsuan Yang\\
UC Merced\\
{\tt\small mhyang@ucmerced.edu}
\and
David C. Noelle\\
UC Merced\\
{\tt\small dnoelle@ucmerced.edu}
}

\maketitle

\ifwacvfinal\thispagestyle{empty}\fi

\begin{abstract}
Deep Convolutional Neural Networks (CNNs) have been repeatedly proven
to perform well on image classification tasks. Object detection
methods, however, are still in need of significant improvements.  In
this paper, we propose a new framework called Ventral-Dorsal Networks
(VDNets) which is inspired by the structure of the human visual
system. Roughly, the visual input signal is analyzed along two
separate neural streams, one in the temporal lobe and the other in the
parietal lobe. The coarse functional distinction between these streams
is between object recognition --- the ``what'' of the signal -- and
extracting location related information --- the ``where'' of the
signal. The ventral pathway from primary visual cortex, entering the
temporal lobe, is dominated by ``what'' information, while the dorsal
pathway, into the parietal lobe, is dominated by ``where''
information. Inspired by this structure, we propose the integration of
a ``Ventral Network'' and a ``Dorsal Network'', which are
complementary. Information about object identity can guide
localization, and location information can guide attention to relevant
image regions, improving object recognition. This new dual network
framework sharpens the focus of object detection. Our experimental
results reveal that the proposed method outperforms state-of-the-art
object detection approaches on PASCAL VOC 2007 by $8\%$ (mAP) and
PASCAL VOC 2012 by $3\%$ (mAP). Moreover, a comparison of techniques
on Yearbook images displays substantial qualitative and quantitative
benefits of VDNet.
\end{abstract}


\section{Introduction}

\begin{figure}[t]
 \centering
\includegraphics[width=0.75\linewidth]{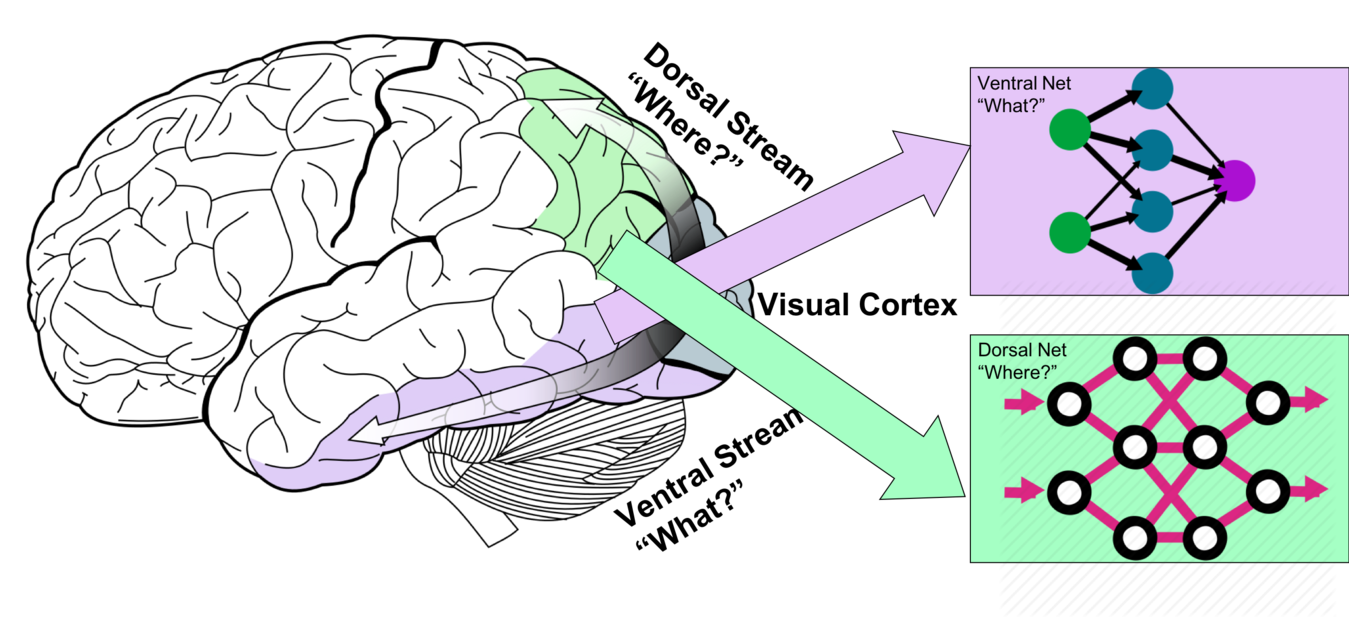}
 \caption{Primary visual cortex and two processing streams. The
   ``ventral stream'' projects into the temporal lobe, and the
   ``dorsal stream'' extends into the parietal lobe. These interacting
   pathways inspire the use of a ``Ventral Net'' and a ``Dorsal
   Net''.} 
 \label{f:brain}
\end{figure}

\textbf{Focus matters.} In order to accurately detect objects in an
image, we need to know which parts of the image are important and then
focus on those parts to find objects of interest. Many approaches to
\emph{selective attention} in artificial neural networks are
computationally expensive, however, limiting their utility for online
applications.

In recent years, deep Convolutional Neural Networks (CNNs) have been
shown to be effective at image classification, accurately performing
object recognition even with thousands of object classes when trained
on a sufficiently rich dataset of labeled
images~\cite{alexnet,vgg16,seNet,denseNet}. One advantage of CNNs is
their ability to learn complete functional mappings from image pixels
to object categories, without any need for the extraction of
hand-engineered image features~\cite{overfeat}. To facilitate
learning through stochastic gradient descent, CNNs embody complex
nonlinear functions which are approximately (due to
activation functions like Relu) differentiable with respect to network
parameters.

Image classification is only one of the core problems of computer
vision, however. Beyond object
recognition~\cite{vgg16,seNet,denseNet}, there are applications for
such capabilities as semantic
segmentation~\cite{semanticSeg1,semanticSeg2,maskrcnn2017}, image
captioning~\cite{caption1,caption2,showAttendTell}, and object
detection~\cite{fastrcnn,fasterrcnn,yolo,yolo9000}. The last of these
involves locating and classifying all of the relevant objects in an
image. This is a challenging problem that has received a good deal of
attention~\cite{fasterrcnn,fastrcnn,rcnn,yolo}. Since there is rarely
\emph{a priori} information about where objects are located in an
image, most approaches to object detection conduct search
over image regions, seeking objects of interest with different sizes
and at different scales. For example, ``region proposal'' frameworks,
like Faster-RCNN~\cite{fasterrcnn}, need to pass a large number of
candidate image regions through a deep network in order to determine
which parts of the image contain the most information concerning
objects of interest. An alternative approach involves one shot
detectors, like Single Shot Detectors (SSD)~\cite{ssd} and You Only
Look Once (YOLO)~\cite{yolo}. These use networks to examine all parts
of the image via a tiling mechanism. For example, YOLO conducts a
search over potential combinations of tiles. In a sense, most
off-the-shelf object detection algorithms distribute attention to all
parts of the image equally, since there is no prior information
concerning where objects of interest might be found.

In contrast, there is much evidence that the human visual system is
equipped with a useful spatial \textbf{selective attention} capability
that guides processing to relevant parts of the scene while
effectively ignoring irrelevant parts~\cite{selAten1995}. Some theories of spatial attention focus on
divergent but interacting neural pathways in the brain's visual
system. Projecting from primary visual cortex in the occipital lobe,
two neural streams emerge, as shown in Figure~\ref{f:brain}. One
stream, extending ventrally into the temporal lobe, appears to largely
encode \emph{what} is in the scene, recognizing objects. The other
stream, passing dorsally into the parietal lobe, approximately
captures information about \emph{where} objects are located. Some
computational neuroscience accounts characterize spatial attention as
naturally arising from interactions between these two neural
pathways~\cite{ccn}. Inspired
by this theory of attention, we propose a novel object detection
framework that integrates two networks that reflect the two pathways
of the human visual system. The \emph{Ventral Net} uses object
classification information to guide attention to relevant image
regions, and the \emph{Dorsal Net} uses this attentional information
to accurately localize and identify objects in the scene. Together,
these two components form a \emph{Ventral-Dorsal Neural Network}
(VDNet). 

Our object detection framework can be seen as arising from the
marriage of \textbf{attention based object
detection}~\cite{top-down-Sal,learning-hier} and \textbf{supervised
object detection}~\cite{yolo,fasterrcnn}. The general approach is to
use ideas from attention based object detection to quickly identify
parts of the image that are irrelevant, with regard to objects of
interest, using this information to guide selective
attention. Focusing further supervised processing on important image
regions is expected to both allow for the allocation of computational
resources in a more informed manner and produce more accurate object
detection by removing potentially distracting background pixels. Our
selective attention process is based on \emph{Sailency Detection} in
activation maps, such as Class Activation Maps~\cite{cam2016}. Guided
by the resulting attentional information, we use supervised object
detection to achieve high object detection performance. Thus, this
framework can be seen as a combination of a top-down attention based
model and the bottom-up detection based approach. As far as we know,
the proposed approach is novel. Moreover, our method performs
favorably against state-of-the-art object detection approaches by a
large margin. In overview, we make these contributions: 
\begin{itemize}
\item Our approach uses a top-down saliency analysis to identify
      irrelevant image regions, introducing a selective attention
      mechanism that masks out noise and unimportant background
      information in the image.
\item Our measure of top-down saliency involves a sensitivity analysis
      of a trained object classification network with regard to the
      Gestalt Total (GT) activation produced by the network.
\item Our method performs very favorably in comparison to
      state-of-the-art approaches, demonstrating an improvement of up
      to 8\% (mAP) in comparsion to the best of those approaches.
\end{itemize} 

The rest of this paper is organized as follows. Section~\ref{s:rw}
reviews related work. In Section~\ref{s:pm}, our proposed framework
for object detection is described. Comparisons of the performance of
VDNet to that of state-of-the-art methods are reported in
Section~\ref{s:exp}. Final conclusions are offered in
Section~\ref{s:conclusion}.


\begin{figure*}[t]
  \centering
    \includegraphics[width=0.8\linewidth]{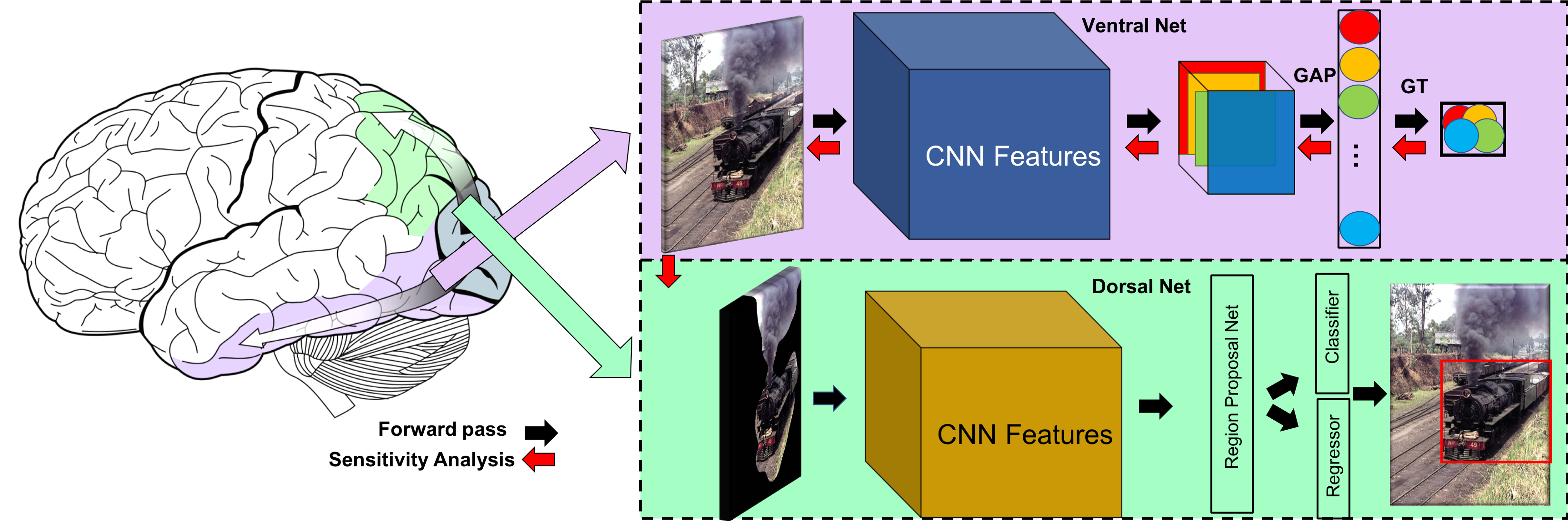}
  \caption{The Ventral-Dorsal Network (VDNet) for Object
           Detection. The Ventral Net filters out irrelevant parts of
           the image based on a sensitivity analysis of the Gestalt
           Total (GT) activation. The Dorsal Net then searches for 
           objects of interest in the remaining regions.}
  \label{f:vdnet}
\end{figure*}
 
\section{Related Work}
\label{s:rw}

State-of-the-art object detection approaches are based on deep CNNs
\cite{fasterrcnn,yolo,yolo9000,maskrcnn2017}. Leading object detection
algorithms can be roughly divided into two categories: attention based
object detection \& supervised object detection.

\subsection{Attention Based Object Detection}

Attention based object detection methods depend on a set of training
images with associated class labels but \emph{without} any
annotations, such as bounding boxes, indicating the locations of
objects. The lack of ground truth bounding boxes is a substantial
benefit of this approach, since manually obtaining such information is
costly.

One object detection approach of this kind is the Class Activation Map
(CAM) method~\cite{cam2016}. This approach is grounded in the
observation that the fully connected layers that appear near the
output of typical CNNs largely discard spatial information. To
compensate for this, the last convolutional layer is scaled up to the
size of the original image, and Global Average Pooling (GAP) is
applied to the result. A linear transformation from the resulting
reduced representation to class labels is learned. The learned weights
to a given class output are taken as indicating the relative
importance of different filters for identifying objects of that
class. For a given image, the individual filter activation patterns in
the upscaled convolutional layer are entered into a weighted sum,
using the linear transformation weights for a class of interest. The
result of this sum is a Class Activation Map that reveals image
regions associated with the target class.

The object detection success of the CAM method has been demonstrated,
but it has also inspired alternative approaches. The work of Selvaraju
et al.~\cite{gradcam2016} suggested that Class Activation Maps could
be extracted from standard image classification networks without any
modifications to the network architecture and additional training to
learn filter weights. The proposed Grad-CAM method computes the
gradients of output labels with respect to the last convolutional
layer, and these gradients are aggregated to produce the filter
weights needed for CAM generation. This is an excellent example of
saliency based approaches that interpret trained deep CNNs, with
others also reported in the
literature~\cite{scale-transfer,progressive,W2F}.

As previously noted, attention based object detection methods benefit
from their lack of dependence on bounding box annotations on training
images. They also tend to be faster than supervised object detection
approaches, producing results by interpreting the internal weights and
activation maps of an image classification CNN. However, these methods
have been found to be less accurate than supervised object detection
techniques.
\vspace{-0.4 mm}
\subsection{Supervised Object Detection}

Supervised object detection approaches require training data that
include both class labels and tight bounding box annotations for each
object of interest. Explicitly training on ground truth bounding boxes
tends to make these approaches more accurate than weakly supervised
methods. These approaches tend to be computationally expensive,
however, due to the need to search through the space of image regions,
processing each region with a deep CNN. Tractability is sought by
reducing the number of image regions considered, selecting from the
space of all possibe regions in an informed manner. Methods vary in
how the search over regions is constrained.

Some algorithms use a \textbf{region proposal based framework}. A deep
CNN is trained to produce both a classification output and location
bounding box coordinates, given an input image. Object detection is
performed by considering a variety of rectangular regions in the
image, making use of the CNN output when only the content in a given
region is presented as input to the network. Importantly, rather than
consider all possible regions, the technique depends on a \emph{region
proposal algorithm} to identify the image regions to be processed by
the CNN. The region proposal method could be either an external
algorithm like Selective Search~\cite{selective-search} or it could be
an internal component of the network, as done in
Faster-RCNN~\cite{fasterrcnn}. The most efficient object detection
methods of this kind are R-CNN~\cite{rcnn}, Fast-RCNN~\cite{fastrcnn},
Faster-RCNN~\cite{fasterrcnn}, and
Mask-RCNN~\cite{maskrcnn2017}. Approaches in this framework tend to be
quite accurate, but they face a number of challenges beyond issues of
speed. For example, in an effort to propose regions containing objects
of one of the known classes, it is common to base region proposals on
information appearing late in the network, such as the last
convolutional layer. The lack of high resolution spatial information
late in the network makes it difficult to detect small objects using
this approach. There are a number of research projects that aim to
address this issue by combining low level features and high level ones
in various ways~\cite{scale-transfer,feature-pyramid}.

Rather than incorporating a region proposal mechanism, some supervised
methods perform \textbf{object detection in one feed-forward pass}. A
prominent method of this kind is YOLO, as well as its
extensions~\cite{yolo,yolo9000}. In this approach, the image is
divided into tiles, and each tile is annotated with anchor boxes of
various sizes, proposing relevant regions. The resulting information,
along with the image tiles, are processed by a deep network in a
single pass in order to find all objects of interest. While this
technique is less accurate than region proposal approaches like
Faster-RCNN, it is much faster, increasing its utility for online
applications.

A comparison of these two general approaches to object detection
displays a clear trade-off between accuracy and computational cost
(speed). This gives rise to the question of whether this trade-off can
be avoided, in some way. In this paper, we propose combining attention
based object detection and supervised object detection in a manner
that leverages the best features of both. Inspired by the human visual
system and its interacting ventral and dorsal streams, we propose
using a fast attention based approach to mask out irrelevant parts of
the original image. The resulting selective attention greatly reduces
the space of image regions to be considered by a subsequent supervised
network.

\section {Method}
\label{s:pm}

\subsection{Overview}

Human vision regularly employs selective attention. Without
introspective awareness of the process, our brains appear to quickly
focus visual processing on important parts of a scene, apparently
masking out unimportant parts. Our proposed object detection framework
is inspired by this phenomenon. It uses a kind of selective attention
to quickly eliminate distractions, backgrounds, and irrelevant areas
from the image before object detection is performed, providing a
sharper focus to the process. VDNet incorporates two distinct
networks: one for quickly guiding selective attention and a second for
classifying and localizing objects of interest.

The \emph{Ventral Net} is used to determine how to apply selective
attention to the image. Ventral Net is a deep CNN that first extracts
convolutional features from the input image and then aggregates the
results into a \emph{Gestalt Total} (GT) output. A sensitivity
analysis is then performed, identifying those pixels in the original
image that have the greatest immediate influence on the GT value. A
simple smoothing operation translates the clouds of salient pixels
into relevant regions in the image. The result is a guide for
selective attention in the original image space. It provides an
indication of which parts of the image are important and which are
irrelevant.

The \emph{Dorsal Net} performs supervised object detection, but it
does so on a modified version of the image in which irrelevant
regions, as determined by the \emph{Ventral Net}, have been simply
masked out. Selective attention speeds the region proposal process by
restricting consideration to unmasked portions of the image. The
result is class labels and bounding boxes for the objects of interest
in the image.

In the remainder of this section, our new framework is described in
detail. VDNet is illustrated in Figure~\ref{f:vdnet}. The following
notation is used: $X\in \calR^{m \times n \times c}$ is the image
input of the system, where $m$, $n$, and $c$ are the width, height,
and number of channels, respectively; an associated ground truth class
label is $z$, and $b \in \calR^{4}$ is the associated ground truth
bounding box coordinates of the object in the input image.

\subsection{Ventral Net}

Ventral Net is responsible for guiding selective attention. It is
built upon the convolutional layers of a deep CNN that has been
trained to perform image classification. The last convolutional layer
is used to calculate the Gestalt Total (GT) activation for the current
input.

For a given image, let $f_k(x,y)$ denote the activation of filter $k$
in the last convolutional layer at spatial location $(x,y)$. For
filter $k$, the Global Average Pooling (GAP) value is defined as:
\begin{equation}
\label{e:CAM}
F^k = \sum_{x,y} f_k(x,y)
\end{equation}
The GT simply aggregates these values across all of the filters:
\begin{equation}
\label{e:GT}
GT = \sum_{k} F^k
\end{equation}
While this simple scalar value might seem to carry little useful
information, it provides us with a simple way to identify the source
pixels that have the greatest influence on the final convolutional
layer activity, in a local sense. These pixels are identified through
a \emph{sensitivity analysis}, measuring the degree to which changes
in an input pixel affect the GT. This is easily computed as follows:
\begin{equation}
S = \left. {\frac{\partial \ GT}{\partial \ X}}\ \right|_{X=I_i}
\label{eq:sensitivity}
\end{equation}
where $X$ is the network input and $I_i \in \calR^{m \times n \times
c}$ is the $i^{th}$ image in the dataset. The result of this
sensitivity analysis is $S \in \calR^{m \times n \times c}$. Since we are interested in all the important pixels in the image; therefore, we work with the absolute values of $S$. Just as
methods for stochastic gradient descent can involve the calculation of
gradients of the output error, for a given input, with regard to
network weights using a single forward pass and a single backward pass
through the network, the value of $S$ can be quickly calculated for
each image in the dataset.

Derivatives are calculated for all of the inputs to the network, which
typically include three channels per pixel (i.e., RGB). Since the
purpose of Ventral Net is to guide spatial attention, a single measure
of relevance for each pixel location, not each channel, is
needed. Thus, some way to aggregate across channels is needed. We have
considered two different methods of aggregation. The first involves
averaging derivative values across channels: 
\begin{equation}
\hat{S}_{x,y}= \frac{1}{k}\sum_k S_{x,y,k}
\label{eq:aggregation_channels1}
\end{equation}   
where $k$ is the number of channels in the input image (i.e., $3$ for
RGB) and $\hat{S}$ is the $\calR^{m \times n}$ result of aggregating
derivatives. An alternative aggregation method is to use the maximum
derivative across channels:
\begin{equation}
\hat{S}_{x,y} = \max_{k} (S_{x,y,k})
\label{eq:aggregation_channels2}
\end{equation}
The resulting $\hat{S}$ provides a measure of relevance at the
individual pixel level. (The evaluation experiments reported in this
paper all used the averaging approach to aggregation.) In order to
translate this information into larger regions of relevance, we smooth
$\hat{S}$ by convolving it with a Gaussian filter, resulting in
$\tilde{S}$.
To extract distinct regions from the resulting smoothed attention map, 
pixels need to be classified as relevant or irrelevant. For
simplicity, this is done by setting to zero any value in $\tilde{S}$
below the mean over all pixels and setting the other values in
$\tilde{S}$ to one. The resulting binary mask is duplicated across
the number of channels in the original image, aligning the
dimensionality of the mask and the image.

The mask is applied to the original image through simple element-wise
multiplication. Formally:
\begin{equation}
\hat{I} = I \odot \tilde{S}
\label{eq:mask}
\end{equation}
where $I$ is the original image and $\hat{I}$ is the masked
image. This modified image then becomes the input to Dorsal Net.

\subsection{Dorsal Net}

Reducing the original image, $I$, to one in which irrelevant regions
are masked out, $\hat{I}$, substantially reduces the space of
candidate regions to consider during object detection. In Dorsal Net,
the masked image is provided as input to a deep CNN trained to propose
regions of interest with anchor boxes, process the contents of those
regions, and output both class labels and bounding box
coordinates. This is done using the methods of
Faster-RCNN~\cite{fasterrcnn}. Dorsal Net is trained using a dataset
of images that are annotated with both ground truth class labels and
ground truth bounding boxes. Network parameters are selected so as to
minimize a combination of the classification loss and the regression
loss arising from the output of bounding box coordinates. The complete
objective function is:
\begin{equation}
\begin{split}
L(p_i,t_i) = \frac{1}{N_{cls}}\sum_i L_{cls}(p_i,p_i^{\ast}) \\
+\lambda \frac{1}{N_{reg}} \sum_i p_i^{\ast} L_{reg}(t_i,t_i^{\ast}).
\end{split}
\label{eq:objFunction}
\end{equation}
where $i$ is the index of an anchor box appearing in the current
training mini-batch and $p_i$ is the predicted probability of anchor
$i$ containing an object of interest. The ground truth label
$p_i^{\ast}$ is $1$ if anchor $i$ is positive for object presence and
it is $0$ otherwise. The predicted bounding box is captured by the 4
element vector $t_i$, and $t_i^{\ast}$ contains the coordinates of the
ground truth bounding box associated with a positive anchor. The two
components of the loss function are normalized by $N_{cls}$ and
$N_{reg}$, and they are weighted by a balancing parameter,
$\lambda$. In our current implementation, the classification loss term
is normalized by the mini-batch size (i.e., $N_{cls} = 32$) and the
bounding box regression loss term is normalized by the number of
anchor locations (i.e., $N_{reg} \approx 2,400$). We set $\lambda =
10$, making the two loss terms roughly equally weighted.

It is worth noting that our general approach could easily support the
implementation of Dorsal Net using a wide variety of alternative
algorithms. The only requirement is that the object detection
algorithm must be able to accept masked input images. For the results
presented in this paper, we have used a leading region proposal based
approach due to the high accuracy values reported for these methods in
the literature. Having Ventral Net reduce the number of proposed
regions is expected to speed the object detection process, and it may
also improve accuracy by removing from consideration irrelevant
portions of the image.

The general architecture of VDNet appears in
Figure~\ref{f:vdnet}. Pseudocode for the proposed VDNet method is
provided in Algorithm~\ref{alg:algorithm1}. An experimental evaluation
of VDNet appears next.

\begin{algorithm}[t]
\caption{Ventral-Dorsal Net Pseudocode}
\label{alg:algorithm1}
\begin{algorithmic}[1]
\State \textbf{Input}: $\I,z,b$, a labeled RGB image
\State \textbf{Output}: $\I$, an RGB image with labeled bounding boxes
around objects of interest
\Procedure{VentralNet}{$\I$}
\State Forward pass $\I$ though the convolutional layers to get $F$.
\State Compute GT based on Equations~\ref{e:CAM} and \ref{e:GT}.
\State Calculate the sensitivity of GT with respect to the input via
       Equation~\ref{eq:sensitivity}.
\State Aggregate the sensitivity maps for spatial locations using
       Equation~\ref{eq:aggregation_channels1} or
       Equation~\ref{eq:aggregation_channels2}.
\State Smooth the result by convolving the aggregated map with a
       Gaussian filter. 
       Threshold the smoothed map to produce a binary mask.
\State Mask the input image using Equation~\ref{eq:mask}, producing
       $\hat{I}$. 
\State \textbf{Return} $\hat{I}$
\EndProcedure

\Procedure{DorsalNet}{$\hat{I}$,$z$,$b$}
\State Forward pass $\hat{I}$ though a pretrained network to obtain
       the convolution features.
\State Run the region proposal network using the resulting feature map.
\State Pool regions of interest (ROIs).
\State Pass pooled ROI to the fully connected layers.
\State \textbf{Return} $\I$ with bounding box around the objects of interest
\EndProcedure
\end{algorithmic}
\end{algorithm}

\begin{figure*}[t]
  \centering
  \begin{tabular}{c@{\hspace{2ex}}c@{\hspace{2ex}}c@{\hspace{2ex}}c@{\hspace{2ex}}c@{\hspace{2ex}}c@{\hspace{2ex}}c@{\hspace{2ex}}c@{\hspace{2ex}}}
    \includegraphics*[width=1.65cm,height=1.6cm]{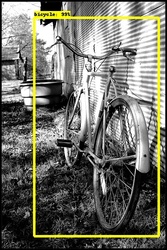}&
    \includegraphics*[width=1.65cm,height=1.6cm]{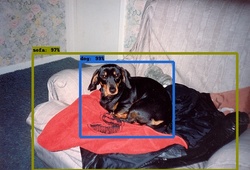}&
    \includegraphics*[width=1.65cm,height=1.6cm]{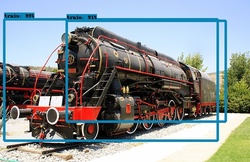}&
    \includegraphics*[width=1.65cm,height=1.6cm]{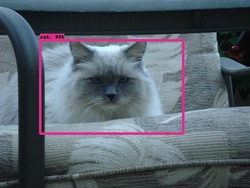}&
    \includegraphics*[width=1.65cm,height=1.6cm]{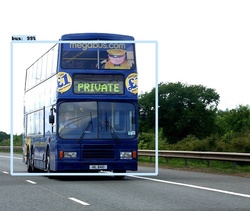}&
    \includegraphics*[width=1.65cm,height=1.6cm]{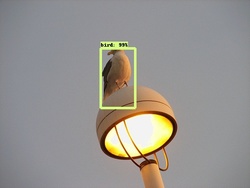}&
    \includegraphics*[width=1.65cm,height=1.6cm]{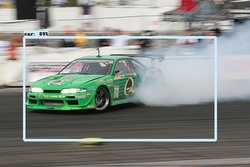}&
    \includegraphics*[width=1.65cm,height=1.6cm]{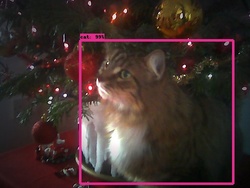}\\
    \includegraphics*[width=1.65cm,height=1.6cm]{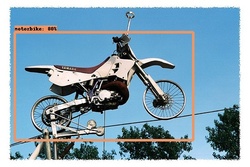}&
    \includegraphics*[width=1.65cm,height=1.6cm]{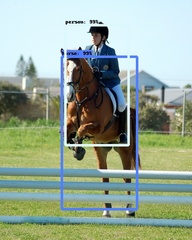}&
    \includegraphics*[width=1.65cm,height=1.6cm]{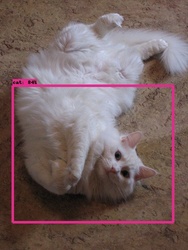}&
    \includegraphics*[width=1.65cm,height=1.6cm]{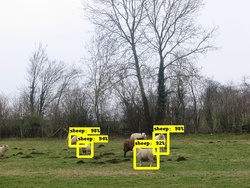}&
    \includegraphics*[width=1.65cm,height=1.6cm]{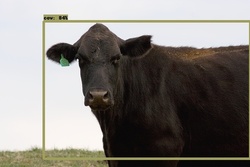}&
    \includegraphics*[width=1.65cm,height=1.6cm]{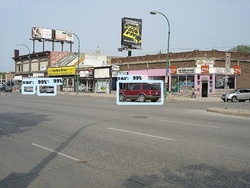}&
    \includegraphics*[width=1.65cm,height=1.6cm]{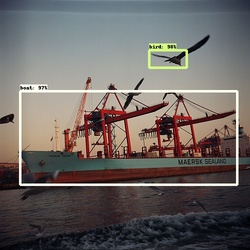}&
    \includegraphics*[width=1.65cm,height=1.6cm]{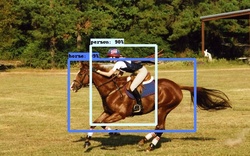}\\
  \end{tabular}
  \caption{VDNet Performance on Some PASCAL VOC 2007 Validation Images}
  \label{f:voc2007-results}
\end{figure*}

\section{Experimental Evaluation}
\label{s:exp}

\subsection{Experiment Design and Implementation}

We evaluated VDNet on PASCAL VOC 2007~\cite{voc2007}, PASCAL VOC
2012~\cite{voc2012}, and on a dataset of Yearbook images. 

The PASCAL VOC 2007 dataset has 20 classes and $9,963$ images which
have been equally split into a training/validation set and a test set.
The PASCAL VOC 2012 dataset contains $54,900$ images from 20 different
categories, and it has been split approximately equally into a
training/validation set and a test set. An additional examined dataset
is a private Yearbook dataset owned by Ancestry.com Operations Inc., used with permission. This dataset has more than 75 million yearbook page
images. We trained a separate model to detect portraits and group
photos in these images.

For PASCAL VOC 2007, we conducted training on the union of the VOC
2007 trainval set and the VOC 2012 trainval set, and we evaluated the
result using the VOC 2007 test set. (This regimen is standard practice
for these datasets.) For PASCAL VOC 2012, we performed training on its
trainval set, and we evaluated the result on its test set. For the
Yearbook dataset, we trained on 1 million images, and we evaluated the
results on 10,000 test set images.

To evaluate performance, we used the standard mean average precision
(mAP) measure. For all datasets we report mAP scores using IoU
thresholds at $0.5$.

For networks with $224 \times 224$ image inputs, using PASCAL VOC, we
trained the model with a mini-batch size of 16 due to GPU memory
constraints. We started the learning rate at $3 \times 10^{-4}$ for the
first $900,000$ epochs. We decreased it to $3 \times 10^{-5}$ until
epoch $1,200,000$. Then, we decreased it to $3 \times 10^{-6}$ until epoch
$2,000,000$. In all cases, we used a momentum optimizer value of
$0.9$.

\begin{table*}[t]
\begin{center}
\scalebox{0.63}{
 \begin{tabular}{c|c|c|cccccccccccccccccccc} 
 \hline
 Method & Network & mAP & areo& bike&bird&boat&bottle&bus&car&cat&chair&cow&table&dog&horse&mbike&person&plant&sheep&sofa&train&tv \\  
 \hline\hline
Faster \cite{fasterrcnn}& VGG& 73.2& 76.5& 79& 70.9& 65.5& 52.1& 83.1& 84.7& 86.4& 52& 81.9& 65.7& 84.8& 84.6& 77.5& 76.7& 38.8& 73.6& 73.9 &83& 72.6 \\ 
ION \cite{bell} & VGG& 75.6& 79.2& 83.1& 77.6& 65.6& 54.9& 85.4& 85.1& 87& 54.4& 80.6& 73.8& 85.3& 82.2& 82.2& 74.4& 47.1& 75.8& 72.7& 84.2& 80.4\\
Faster \cite{resNet}& Residual-101& 76.4& 79.8& 80.7& 76.2& 68.3& 55.9& 85.1& 85.3& 89.8& 56.7& 87.8& 69.4& 88.3& 88.9& 80.9& 78.4& 41.7& 78.6& 79.8& 85.3& 72\\
MR-CNN \cite{multi-region}& VGG& 78.2& 80.3& 84.1& 78.5& 70.8& 68.5& 88& 85.9& 87.8& 60.3& 85.2& 73.7& 87.2& 86.5& 85& 76.4& 48.5& 76.3& 75.5& 85& 81\\
R-FCN \cite{region-based}& Residual-101& 80.5& 79.9& 87.2& 81.5& 72& 69.8& 86.8& 88.5& 89.8& 67& 88.1& 74.5& 89.8& 90.6& 79.9& 81.2& 53.7& 81.8& 81.5& 85.9& 79.9\\
\hline
SSD300 \cite{ssd} &VGG& 77.5& 79.5& 83.9& 76& 69.6& 50.5& 87& 85.7& 88.1& 60.3& 81.5& 77& 86.1& 87.5& 83.9& 79.4& 52.3& 77.9& 79.5& 87.6& 76.8 \\
SSD512 \cite{ssd}& VGG& 79.5& 84.8& 85.1& 81.5& 73& 57.8& 87.8& 88.3& 87.4& 63.5& 85.4& 73.2& 86.2& 86.7& 83.9& 82.5& 55.6& 81.7& 79& 86.6& 80\\
\hline
DSSD321 \cite{dssd}& Residual-101& 78.6& 81.9& 84.9& 80.5& 68.4& 53.9& 85.6& 86.2& 88.9& 61.1& 83.5& 78.7& 86.7& 88.7& 86.7& 79.7& 51.7& 78& 80.9& 87.2& 79.4\\
DSSD513 \cite{dssd}& Residual-101& 81.5& 86.6& 86.2& 82.6& 74.9& 62.5& 89& 88.7& 88.8& 65.2& 87& 78.7& 88.2& 89& 87.5& 83.7& 51.1& 86.3& 81.6& 85.7& 83.7\\
\hline
STDN300 \cite{scale-transfer} &DenseNet-169&78.1& 81.1& 86.9& 76.4& 69.2& 52.4& 87.7& 84.2& 88.3& 60.2& 81.3& 77.6& 86.6& 88.9& 87.8& 76.8& 51.8& 78.4& 81.3& 87.5& 77.8\\
STDN321 \cite{scale-transfer}& DenseNet-169& 79.3& 81.2& 88.3& 78.1& 72.2& 54.3& 87.6& 86.5& 88.8& 63.5& 83.2& 79.4& 86.1& 89.3& 88.0& 77.3& 52.5& 80.3& 80.8& 86.3& 82.1\\
STDN513 \cite{scale-transfer}& DenseNet-169& 80.9& 86.1& 89.3& 79.5& 74.3& 61.9& 88.5& 88.3& 89.4& 67.4& 86.5& 79.5& 86.4& 89.2& 88.5& 79.3& 53.0& 77.9& 81.4& 86.6& 85.5\\
\hline
VDNet& Resnet-101&\textbf{86.2}& 95.8& 98.1& 98.4& 65.1& 94.6& 90.1& 96.2& 71.7& 72.3& 54.6& 97.9& 95.6& 89.2& 90.1& 93.2& 69.1& 89.2& 82.1& 93.4.6& 74.0\\
\end{tabular}}
\end{center}
\caption{PASCAL VOC 2007 Test Detection Results. Note that the minimum
         dimension of the input image for Faster and R-FCN is 600, and
         the speed is less than 10 frames per second. SSD300 indicates
         the input image dimension of SSD is $300 \times 300$. Large
         input sizes can lead to better results, but this increases
         running times. All models were trained on the union of the
         trainval set from VOC 2007 and VOC 2012 and tested on the VOC
         2007 test set. \label{t:voc2007}} 
\end{table*}

\begin{table*}[t]
\begin{center}
\scalebox{0.63}{
 \begin{tabular}{c|c|cccccccccccccccccccc} 
 \hline
Method & mAP& aero& bike& bird& boat& bottle& bus& car& cat& chair& cow& table& dog& horse& mbike& person& plant& sheep& sofa& train& tv \\
 \hline\hline
HyperNet-VGG \cite{hyperNet}& 71.4& 84.2& 78.5& 73.6& 55.6& 53.7& 78.7& 79.8& 87.7& 49.6& 74.97 &52.1& 86.0& 81.7& 83.3& 81.8& 48.6& 73.5& 59.4& 79.9& 65.7\\
HyperNet-SP \cite{hyperNet}& 71.3& 84.1& 78.3& 73.3& 55.5& 53.6& 78.6& 79.6& 87.5& 49.5& 74.9& 52.1& 85.6& 81.6& 83.2& 81.6& 48.4& 73.2& 59.3& 79.7& 65.6\\
Fast R-CNN $+$ YOLO \cite{yolo}& 70.7& 83.4& 78.5& 73.5& 55.8& 43.4& 79.1& 73.1& 89.4& 49.4& 75.5& 57.0& 87.5& 80.9& 81.0& 74.7& 41.8& 71.5& 68.5& 82.1& 67.2 \\
MR-CNN-S-CNN \cite{MR-CNN-S-CNN}& 70.7& 85.0& 79.6& 71.5& 55.3& 57.7& 76.0& 73.9& 84.6& 50.5& 74.3& 61.7& 85.5& 79.9& 81.7& 76.4& 41.0& 69.0& 61.2& 77.7& 72.1 \\
Faster R-CNN \cite{fasterrcnn}& 70.4& 84.9& 79.8& 74.3& 53.9& 49.8& 77.5& 75.9& 88.5& 45.6& 77.1& 55.3& 86.9& 81.7& 80.9& 79.6& 40.1& 72.6& 60.9& 81.2& 61.5 \\
NoC \cite{NoC}& 68.8& 82.8& 79.0& 71.6& 52.3& 53.7& 74.1& 69.0& 84.9& 46.9& 74.3& 53.1& 85.0& 81.3& 79.5& 72.2& 38.9& 72.4& 59.5& 76.7& 68.1\\
VDNet & \textbf{73.2}& 85.1& 82.4& 73.6& 57.7& 61.2& 79.2& 77.1& 85.5& 54.9& 79.8& 61.4& 87.1& 83.6& 81.7& 77.9& 45.6& 74.1& 64.9& 80.3& 73.1\\

\end{tabular}}
\end{center}
\caption{PASCAL VOC 2012 Test Detection Results. Note that the
         performance of VDNet is about $3 \%$ better than baseline
         Faster-RCNN. \label{t:voc2012}} 
\end{table*}


\begin{figure}[b]
  \centering
  \begin{tabular}{c@{\hspace{2ex}}c@{\hspace{2ex}}c@{\hspace{2ex}}c@{\hspace{2ex}}}
    \includegraphics*[width=1.65cm,height=1.6cm]{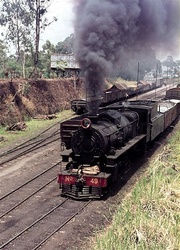}&
    \includegraphics*[width=1.65cm,height=1.6cm]{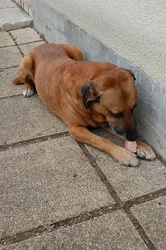}&
    \includegraphics*[width=1.65cm,height=1.6cm]{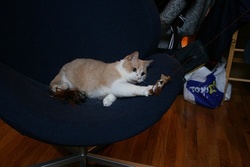}&
    \includegraphics*[width=1.65cm,height=1.6cm]{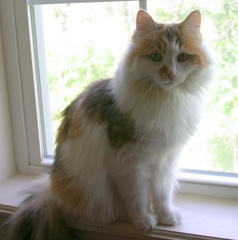}\\
    \includegraphics*[width=1.65cm,height=1.6cm]{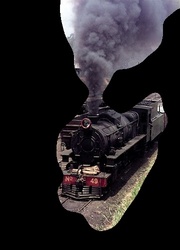}&
    \includegraphics*[width=1.65cm,height=1.6cm]{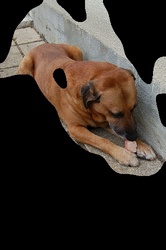}&
    \includegraphics*[width=1.65cm,height=1.6cm]{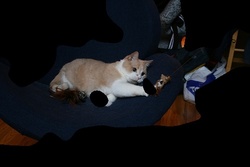}&
    \includegraphics*[width=1.65cm,height=1.6cm]{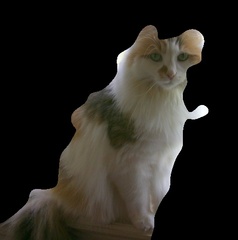}\\

 \end{tabular}
 \caption{Some examples of selective attention image masking based on
          Ventral Net sensitivity analyses. The first row shows
          original images, and the second row displays the results of
          Ventral Net based masking.}
  \label{f:ventralNet}
\end{figure}

\subsection{PASCAL VOC 2007}

\subsubsection{Comparative Performance Results}

The results of the PASCAL VOC 2007 dataset evaluation appear in
Table~\ref{t:voc2007}. For the Ventral Net, we utilized VGG19
pretrained on the ImageNet dataset. We removed the fully connected
layers and softmax calculation from VGG19, and we calculated GT based
on the last convolutional layer. The Ventral Net sensitivity analysis
was performed on the resulting network. No fine tuning of parameters
was done. For the Dorsal Net we used Resnet 101 pretrained on the
ImageNet dataset as the backbone network for object detection. The
model was trained on 8 GPUs hosted by Amazon Web Services (AWS) for
about one week.

We compared our performance results with those reported for a variety
of state-of-the-art approaches to object detection. Our primary
baseline was Faster-RCNN using a Resnet 101 network trained on PASCAL
VOC 2007. As shown in Table~\ref{t:voc2007}, the selective attention
process of our approach (VDNet) resulted in substantially better
performance in comparison to Faster-RCNN and other methods. VDNet
appears to be more accurate at detecting larger objects than smaller
ones, perhaps because the region proposal network based its output on
the last (lowest resolution) convolutional layer. For most of the
object classes, VDNet performed better than other methods by a large
margin. For some classes, however, the selective attention mechanism
sometimes failed to identify the relevant parts of the image,
resulting in poorer performance. Some illustrative test data examples
are provided in Figures~\ref{f:voc2007-results} and~\ref{f:ventralNet}.

\subsubsection{Discussion}

In order to further understand the contributions of the various
components of VDNet, we observed the results of focal modifications to
the system.

We noticed that poor performance could frequently be traced to poor
selective attention masks. By varying parameters, we found that the
variance of the Gaussian filter used to smooth the sensitivity
analysis map played an important role. A poor choice for the variance
could result in highly inappropriate attentional masks. We found that
good performance could be had on the PASCAL datasets by using a
variance value between $25$ and $35$. It is likely, however, that this
value would need to be tuned to the size and kinds of objects to be
detected. 

We considered performing the Ventral Net sensitivity analysis using
the output class labels of the pretrained image classification
network, rather than basing that analysis on the Gestalt Total (GT)
activation of the last convolutional layer. It seems natural to ask
for the set of pixels that contribute most to the recognition of an
object of a particular class. There are a couple of reasons why we
found sensitivity of GT to pixels to be a better measure. First,
focusing on the last convolutional layer allowed us to produce
reasonable sensitivity maps even for object classes novel to the
pretrained classification network. Second, since object detection
often involves scenes containing multiple objects, producing
sensitivity maps based on class outputs would require the aggregation
of sensitivity information for each class that might appear in the
image. In the most general case, this means calculating a separate
sensitivity map for each class, increasing the execution time of the
Ventral Net process by a factor equal to the number of classes.

We performed a cursory examination of the role of network depth in the
Dorsal Net on object detection performance on the PASCAL VOC 2007
dataset. In general, it seems as if increasing network depth increases
object detection accuracy. This observation was based on comparing
three different implementations of the Dorsal Net: Inception, Resnet
50, and Resnet 101. Accuracy was much better for the deeper networks,
as shown in Table~\ref{t:diffrentNet}.
\begin{table}[h]
\begin{center}
 \begin{tabular}{c|c|c} 
 \hline
VDNet Component & Deep Network & mAP \\
 \hline
Dorsal Net & Inception & 63.1 \\
Dorsal Net & ResNet50 & 71.6 \\
Dorsal Net & ResNet101 & 86.2 \\
\hline
\end{tabular}
\end{center}
\caption{PASCAL VOC 2007 Test Results for Different Network
         Architectures \label{t:diffrentNet}}
\end{table}

\subsection{PASCAL VOC 2012}

We also measured VDNet performance on the PASCAL VOC 2012 dataset. The
Ventral Net consisted of VGG19, with features for calculating GT
extracted from the last convolutional layer. The Dorsal Net was
initialized with parameters previously learned for the PASCAL VOC 2007
evaluation, but further training was done. For the additional
training, the learning rate was initialized to $3 \times 10^{-4}$ for
$900,000$ epochs, and then it was reduced to $3 \times 10^{-5}$ until
reaching epoch $1,200,000$. The learning rate was further reduced to
$3 \times 10^{-6}$ until reaching $3,000,000$ epochs. The whole
training process took about 14 days on 8 GPUs hosted by AWS. This
resulted in a VDNet that produced comparable or better performance
that state-of-the-art methods. Performance results for PASCAL VOC 2012
are shown in Table~\ref{t:voc2012}.

\subsection{Yearbook Dataset}

We performed additional evaluation experiments using a dataset of
grayscale Yearbook page images, testing VDNet in a somewhat different
application domain. This dataset is not publicly available, but was
provided by Ancestry.com Operations Inc, which possesses more than 75 million
yearbook page images. Pictures on these pages appear at a variety of
different angles, with different shapes, scales, and sizes. The task
was to automatically detect and crop the portraits and the group
photos from the pages. 

Because of the differences in image properties, it did not make sense
to use a network pretrained on the ImageNet dataset as the Ventral
Net. Instead, we produced a classification network appropriate for
this task. Our architecture included two convolutional layers, with 32
and 64 filters, respectively. The last convolutional layer was
followed by max pooling and GAP, feeding a fully connected layer with
Relu units followed by a softmax operation. Classification was done to
recognize group photos and portraits in $224 \times 224$ image
patches. A batch size of 32 was used, and training was conducted for
$1,000$ epochs. No regularization (e.g., weight decay, drop out) was
used. The learning rate was initialized to $10^{-1}$, and it decayed
exponentially every $1,000$ epochs. The Dorsal Net was Resnet 101.

We compared our VDNet to Faster-RCNN and discovered that VDNet
exhibited a $7\%$ improvement in accuracy. 

\section{Conclusion}
\label{s:conclusion}

In this paper, we highlighted the utility of incorporating fast
selective attention mechanisms into object detection algorithms. We
suggested that such mechanisms could potentially speed processing by
guiding the search over image regions, focusing this search in an
informed manner. In addition, we demonstrated that the resulting
removal of distracting irrelevant material can improve object
detection accuracy substantially.

Our approach was inspired by the visual system of the human
brain. Theories of spatial attention that see it as arising from dual
interacting ``what'' and ``where'' visual streams led us to propose a
dual network architecture for object detection. Our Ventral Net
consists of a pretrained image classification network, and a
sensitivity analysis of this network, focusing on the Gestalt Total
(GT) activation of the final convolutional layer, is performed to
provide a fast identification of relevant image regions. The Ventral
Net guides selective attention in the Dorsal Net, restricting the
spatial regions considered by the Dorsal Net as it performs high
accuracy object detection. Our approach, VDNet, integrates attention
based object detection methods with supervised approaches.

The benefits of selective attention, as implemented in VDNet, are
evident in performance results on the PASCAL VOC 2007 and PASCAL VOC
2012 datasets. Evaluation experiments revealed that VDNet displays
greater object detection accuracy than state-of-the-art approaches,
often by a large margin.

There are many opportunities for improving upon VDNet. One noteworthy
weakness was observed in the detection of smaller objects, arising
from the use of later, lower resolution, convolutional layers for the
generation of region proposals. Ongoing research is focusing on
equipping VDNet with network architectures that support learning at
multiple spatial scales. The idea is to perform senstivity analyses in
Ventral Net based on convolutional layers throughout the
classification network, combining these in a manner that integrates
information from smaller spatial scales without being distracted by
more fundamental image features. It is important that the result of
the sensitivity analysis continues to focus on the visual properties
that are closely related to the presence of objects of interest.

Finally, it is worth noting that VDNet has a very general and flexible
structure. The Ventral Net and the Dorsal Net can easily be updated to
incorporate the latest advancements in image classification and object
detection algorithms. In this way, we see VDNet as opening many
avenues for future exploration. \\

\section{Acknowledgment}
This work was started as an internship research project at Ancestry.com Operations Inc. and continued at UC Merced. We would like to thank Ancestry.com Operations Inc. data science division for useful discussions and providing computational GPU resources.\\

{\small
\bibliographystyle{ieee}
\bibliography{egbib}
}

\end{document}